\documentclass[journal,singlespace,twocolumn,10pt]{IEEEtran}
\usepackage{color,amsmath}

\ifCLASSINFOpdf
\else
   \usepackage[dvips]{graphicx}
\fi
\usepackage{url}

\usepackage{graphicx}

\begin{document}

\title{Integrating global spatial features in CNN based Hyperspectral/SAR imagery classification.}

\author{Fan Zhang, \IEEEmembership{Senior Member, IEEE}, MinChao Yan, Chen Hu, Jun Ni*, \IEEEmembership{Student Member, IEEE}, Fei Ma,
\thanks{This work was supported in part by the National Natural Science Foundation of China under Grant 61871413, 61801015. (\emph{Corresponding author: Jun Ni.})}
\thanks{F. Zhang, M Yan, F. Ma and J. Ni are with the College of Information Science and Technology, Beijing University of Chemical Technology, Beijing 100029, P.R.China.}
\thanks{C. Hu is with Sugon Information Industry Co., Ltd, Beijing 100193, P.R.China.}
}


\markboth{This paper is submitted to letter}
{Shell \MakeLowercase{\textit{et al.}}: Bare Demo of IEEEtran.cls for IEEE Journals}
\maketitle

\begin{abstract}
\textcolor{black}{The land cover classification has played an important role in remote sensing because it can intelligently identify things in one huge remote sensing image to reduce the work of humans. However, a lot of classification methods are designed based on the pixel feature or limited spatial feature of the remote sensing image, which limits the classification accuracy and universality of their methods.  This paper proposed a novel method to take into the information of remote sensing image, i.e., geographic latitude-longitude information. In addition, a dual-branch convolutional neural network (CNN) classification method is designed in combination with the global information to mine the pixel features of the image. Then, the features of the two neural networks are fused with another fully neural network to realize the classification of remote sensing images. Finally, two remote sensing images are used to verify the effectiveness of our method, including hyperspectral imaging (HSI) and polarimetric synthetic aperture radar (PolSAR) imagery. The result of the proposed method is superior to the traditional single-channel convolutional neural network.}
\end{abstract}


\IEEEpeerreviewmaketitle

\section{Introduction}

\IEEEPARstart{R}{emote} \textcolor{black}{sensing images have been used in various fields of civil and military applications, and land cover classification is one of the most important applications of remote sensing images. A land use object can contain many different land cover elements to form complex structures, and a specific land cover type can be a apart of different land use objects \cite{Douglas2015Mapping, TONG2020111322}. With those abundant spatial features and image information, classification methods can distinguish the types of ground objects with high accuracy. In the pixel-based classification, the classification process is to classify feedback signals according to the different absorption rate and reflectivity of surface materials \cite{Bhosle2019}.}


Recently, the neural network (NN) has achieved great success in many visual tasks such as image recognition \cite{Ma2015Hyperspectral}, object feature extraction \cite{zhang2019sar}, semantic segmentation, and so on.  Owing to the powerful feature extraction ability of the neural network, it has also generated widespread interest in remote sensing classification \cite{8921180}. The convolutional neural network (CNN) can extract more abstract and invariant features in remote sensing images, and has proven its superior classification performance \cite{Okeke}.  As a result, researchers began to focus on the development of neural networks in the field of land cover classification \cite{zhong2017a}. Obviously, spatial information has a significant impact on image classification. The neighborhood-information of pixels was always introduced to optimize classification results in the past, but this approach does not make use of the global spatial information of the image.


Therefore, image segmentation is widely used in post-processing of remote sensing classification, e.g., Markov Random Field (MRF) \cite{Cao2018Hyperspectral,8127432}. Meanwhile, global information has been used to consider more spatial features in feature extraction and classification \cite{zhongyang2019terrain,wolf2015fast,krahenbuhl2011efficient}. 
Although MRFs and CRFs utilize local nodal interaction in modeling, they will cause excessive smoothness on boundaries. In the dense conditional random field (DenseCRF) method \cite{krahenbuhl2011efficient}, one pixel is connected with all the other pixels to establish an energy function to capture non-local relationships. Although the CRF method is widely used in post-processing of remote sensing classification, a large number of independent parameters limit its practical application. Their method inspired us to propose a novel approach that combines the global features of remote sensing images with the traditional features of a pixel.


In addition, the dual-branch NN method that has proved to be advantageous in remote sensing classification is designed to extract pixel features and the coordinate feature \cite{Chen2014Deep}. In our method, CNN is designed to extract the traditional pixel feature, while the other fully connected network (FCN) \cite{Liu2018An} is intended to excavate the coordinate feature to supplement remote sensing feature. The results of the two branches will be fused by addition, and another fully connected network will be employed to obtain the final decision classification. Compared with the existing land cover classification methods, we made the following contributions:
\begin{enumerate}
    \item The global information, i.e. the geographic latitude-longitude feature, is proposed for the first time to enhance the remote sensing classification.
    \item Aiming at the difference between pixel features and geographic latitude-longitude features, we designed a dual-branch neural network method to extract them respectively to distinguish spatial features and pixel features.


\end{enumerate}

To verify the effectiveness of our method, we used two different remote sensing datasets for experiments, namely hyperspectral image (HSI) and polarimetric synthetic aperture radar (PolSAR) image.

\section{Proposed method}

\begin{figure*}[http]
\centering
\includegraphics[width=15cm]{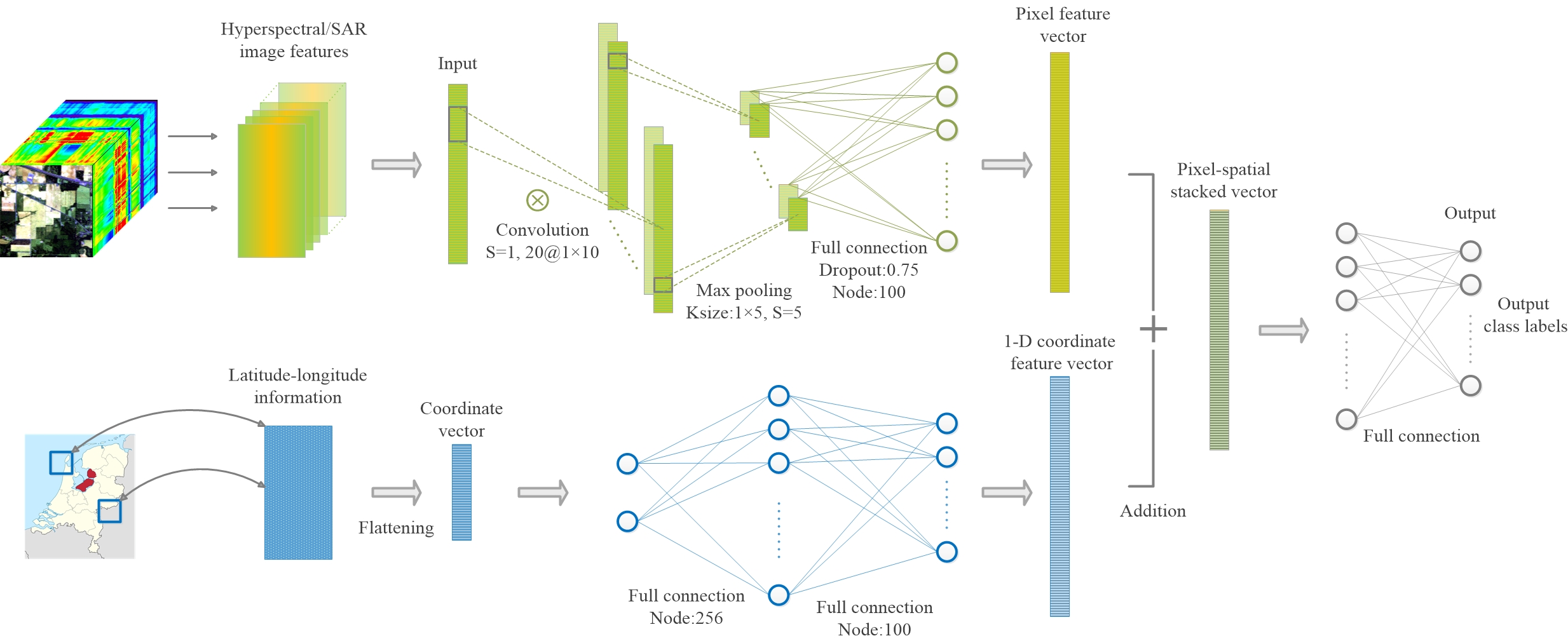}
\caption{The proposed remote sensing image classification framework.}
\label{figure1}
\end{figure*}

The proposed method consists of two frameworks: pixel feature extraction and feature learning of global information. The overall architecture of the remote sensing image classification framework is shown in Figure \ref{figure1}.

The upper-branch neural network consists of 1-dimensional CNN, which takes the pixel feature vector of remote sensing as input and output the extracted pixel feature.

\subsection{The construction of 1D-CNN for pixel feature.}
The structure of CNN can be roughly divided into two parts. The first part is the feature extraction part composed of convolution and pooling operations, and the second part is the classification part aims to map the extracted high-level abstract features to the classification labels. The convolution operation of CNN can be seen as a feature selection of input data with different filters. Through appropriate training, the network structure can learn filter parameters effectively to replace the manual design of features in the traditional feature extraction method and get better characteristics at the same time. The architecture of 1D-CNN in this paper is mainly designed for pixel extraction.


In the 1D-CNN, the relationship between the input and output of the convolution layer can be defined as:

\begin{equation}
\label{equ:1}
V_j = f(\sum\nolimits_{i = 1}^I { {W^1_{ij}}\odot X + {b_j}}),
\end{equation}
where $V_j$ is the output of node $j$, $W^1_{ij}$ is the weight matrix, $\odot$ is the convolution operation, $X$ is the pixel feature vector input of the remote sensing image and $b_j$ is the bias. $f(\cdot)$ is the activation function, which uses $ReLU$. In the convolution layer, the number of nodes is set to 20, the stride is 1, and the size of the convolution kernel is $1\times 10$.

Then, the $V_j$ will be processed by the max-pooling layer and the fully connected layer, which is expressed by
\begin{equation}
\label{equ:2}
O = f(W^2 \cdot V + b),
\end{equation}
where $W^2$ is a weight matrix composed of fully connected layers, $V$ is the result of the pooling layer, and $O$ is the extraction feature of 1D-CNN. In fully connected layer, the dropout rate is 0.75, and the number of node is 100.

\subsection{The design of dual-branch neural network.}
Generally, the probability graph result of the classifier can be fused with the neighborhood-information of pixels by segmentation methods to improve the classification accuracy \cite{zhang2017nearest,xiong2014representation}. Their methods transform the classified probability graph into an energy function, which is established by the pixel probability information and its neighborhood information. However, the local segmentation methods lead to excessive smoothness on boundaries. The dense conditional random field (DenseCRF) method is implemented to capture non-local relationship of a pixel connected to all other pixels \cite{teichmann2018convolutional,8726302}. In the DenseCRF, the energy is defined as


\begin{equation}
\label{equ:2}
E(x_i) = \psi_u(x_i)+  \sum\nolimits_{j = 1, i\neq j}^{C} \psi_p(x_i,x_j),
\end{equation}
where $C$ is the number of pixels. The unary potential $\psi_u(x_i)$ is obtained by the probability graph in classification method, and the pairwise-potential function $\psi_p(x_i,x_j)$ is defined as weighted sum of Gaussian kernels. The entire pairwise potential is given as:
\begin{align}
 \psi_p(x_i,x_j)   & = {w^{\left( 1 \right)}}\underbrace {\exp \left( { - \frac{{{{\left| {{p_i} - {p_j}} \right|}^2}}}{{2\theta _\alpha ^2}} - \frac{{{{\left| {{I_i} - {I_j}} \right|}^2}}}{{2\theta _\beta ^2}}} \right)}_{appearance\;kernel} \\ \nonumber
   & + {w^{\left( 2 \right)}}\underbrace {\exp \left( { - \frac{{{{\left| {{p_i} - {p_j}} \right|}^2}}}{{2\theta _\gamma ^2}}} \right)}_{smoothness\;kernel}
\end{align}
where $I_i$ and $I_j$ are the color vectors; $\theta _\alpha$, $\theta _\beta$, and $\theta _\gamma$ are the control parameters; $w^{(1)}$ and $w^{(2)}$  are the weight parameters; $p_i$ and $p_j$ are the positions. Although the $appearance$ kernel has been widely used in deep learning to solve the classification problem, the $smoothness$ kernel, which based on the spatial coordinates $p_i$ and $p_j$, always be ignored. In fact, global coordinates information is much useful to improve the classification results. Besides, it's true that the pixel in similar positions usually have the same classification in traditional image classification, and vegetation cover of the same latitude and longitude is usually highly correlated in ecological theory \cite{samiappan2016fusion,blanco2010synergistic,wang2017research}. Inspired by their theories, a dual-branch CNN structure that fuses the global feature of remote sensing images with traditional features of the pixel was designed, as shown in Fig. \ref{figure1}.

Since the coordinate information of remote sensing pixels has only two features, a fully connected neural network is used to extract its primary information. In the two layers, the first layer network has 256 nodes to expand the coordinate vector information, and the second layer network has 100 nodes to reduce the feature dimension of the output result of the previous layer network. Finally, the coordinate feature vector is output from the second network.

Then, a dual-branch neural network is designed to fuse the pixel feature vector of 1D-CNN and the coordinate feature vector of FCN. Assuming that $O_1$ is the pixel feature vector of 1D-CNN and $O_2$ is the coordinate feature vector of FCN, and their feature vectors are fused by vector addition

\begin{equation}
\label{equ:3}
O_f=O_1+ O_2.
\end{equation}

Then, the fusion vector $O_f$ is operated in the fully connected network
\begin{equation}
\label{equ:4}
P = f(W\cdot O_f+b),
\end{equation}
where $softmax$ is used for the activate function, $W$ is the weight matrix of FCN, and the final output of our method is $P$ corresponding to the probability of different labels.


In our experiments, $Adam$ is used in the optimizer, $cross-entropy$ is used in the loss function, and the number of maximum epoch is set to 500 .

\section{Experiments}
In this section, two experiments are implemented to verify the effectiveness of the proposed method with two kinds of remote sensing datasets, including hyperspectral image in Indian Pines and PolSAR image in Flevoland respectively. In addition, the normalization processing is carried out on the original dataset before the experiments.

\subsection{Two remote sensing datasets in the experiments.}

\begin{figure}[http]
\centering
\includegraphics[width=8.5cm]{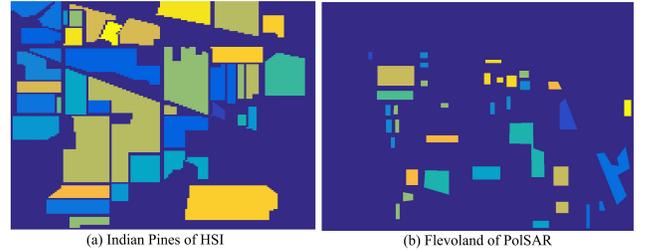}
\caption{The ground truth of two datasets.}
\label{figure3}
\end{figure}

The ground truths of two datasets are shown in the Figure \ref{figure3}. In the first dataset, the size of the Indian Pine is $145\times145$, the wavelength range of the spectrum is 0.4-2.5 microns, and the spatial resolution is 20m. After removing a few poorly performing spectra, 220 spectral channels are retained. The Indian Pines scene contains two-thirds of the agriculture and one-third of the forest or other natural perennial vegetation.

In the second dataset, the size of Flevoland is $750 \times 1024$, which acquired by NASA/JPL AIRSAR system in Flevoland, Netherlands, August 1989 \cite{zhang2017nearest}. In the experiment, 107 features are adopted from different polarimetric descriptors, elements of coherency $\&$ covariance matrix, and the target decomposition theorems.

%
%

\subsection{Experiment results.}
To prove the effectiveness of our method, a 1D-CNN \cite{hu2015deep,ahishali2020multifrequency} and SVM \cite{zhang2017nearest} are also performed using for the comparison experiments. The structure of single-channel CNN is same as the first branch of our proposed method in fig. \ref{figure1}, and the parameters of their networks are consistent. The overall accuracy (OA), average accuracy (AA), and Kappa coefficient are used as the evaluation criteria for classification results.

In the first experiment, available ground truth is designated in 16 categories in HSI. Approximately $5\%$ training samples are randomly selected from the labeled samples and the classification result is shown in Table \ref{tab:1}. 
\begin{table}[h]
\caption{The classification result of Indian Pines in HSI.}
\label{tab:1}
\centering
\begin{tabular}{c|c|c|c|c|c}
\hline
Category&Test   &Training&1D&SVM&Proposed\\
~       &samples&samples &-CNN   &  ~&method\\ \hline
Alfalfa&46&3&8.7&0&97.83\\ \hline
Corn-notill&1428&72&39.92&61.62&90.62\\ \hline
Corn-mintill&830&42&32.53&40.48&89.28\\ \hline
Corn&237&12&21.94&29.96&99.16\\ \hline
Grass-pasture&483&25&54.24&70.81&96.89\\ \hline
Grass-trees&730&37&91.37&90.68&100\\ \hline
Grass-pasture-mowed&28&2&17.86&0&21.43\\ \hline
Hay-windrowed&478&24&95.19&98.12&100\\ \hline
Oats&20&2&20&0&60\\ \hline
Soybean-notill&972&49&42.7&47.84&88.48\\ \hline
Soybean-mintill&2455&123&71.36&85.3&96.95\\ \hline
Soybean-clean&593&30&26.31&41.32&85.67\\ \hline
Wheat&205&11&89.76&92.2&99.51\\ \hline
Woods&1265&64&93.44&92.49&99.92\\ \hline
Buildings-Grass&386&20&31.35&27.72&99.74\\
-Trees-Drives&&&&&\\ \hline
Stone-Steel-Towers&93&5&74.19&79.57&91.4\\ \hline
OA&-&-&60.18&69.31&94.59\\ \hline
AA&-&-&50.68&53.63&88.55\\ \hline
Kappa&-&-&0.54&0.64&0.94\\ \hline
\end{tabular}
\end{table}

Compared with 1D-CNN and SVM, the classification accuracy has been greatly improved in our method. Although the AA, OA and Kappa of SVM are better than those of 1D-CNN, the SVM method has poor classification of limited samples, e.g., Alfalfa, Grass-pasture-mowed and Oats. The classification accuracy of Alfalfa in 1D-CNN is only 8.7\%, while its classification accuracy can reach 97.83\% when the coordinate information is introduced in our method. In addition, the classification accuracy of Corn-notill, Corn-mintill, Corn, Oats, Soybean-clean, and Buildings-Grass can be improved by more than 50\%. Compared with the 1D-CNN, our method can dramatically improve OA, AA and  Kappa to 94.59\%, 88.55\% and 0.94, respectively. 

In the second experiment, 11 different land cover types are marked in the ground truth in PolSAR, and $1\%$ of the labeled samples are selected as the training samples. The classification result is shown in Table \ref{tab:2}.

\begin{table}[h]
\caption{The classification result of Flevoland Pine in PolSAR.}
\label{tab:2}
\centering
\begin{tabular}{c|c|c|c|c|c}
\hline
Category&Test   &Training&1D&SVM&Proposed\\
~       &samples&samples &-CNN   &  ~&method\\ \hline
Stem beans&4121&41&91.87&84.88&97.84\\ \hline
Forest&10109&101&78.73&86.96&99.94\\ \hline
Potato&4848&48&81.37&67.53&98.33\\ \hline
Alfalfa&5132&51&96.34&92.28&99.55\\ \hline
Wheat&14587&145&86.89&89.5&99.81\\ \hline
Bare land&3451&34&94.96&92.96&100\\ \hline
Beet&3977&39&85.77&82.42&92.13\\ \hline
Rapeseed&12469&124&90.2&86.09&99.62\\ \hline
Water&5337&53&86.6&77.85&98.22\\ \hline
Pea&2938&29&87.24&78.66&99.39\\ \hline
Grassland&1219&12&99.26&95&100\\ \hline
OA&-&-&87.46&85.35&98.97\\ \hline
AA&-&-&89.02&84.92&98.62\\ \hline
Kappa&-&-&0.86&0.83&0.99\\ \hline
\end{tabular}
\end{table}

 Although the improvement of classification accuracy in the second experiment is not as obvious as in the first experiment, the advantage of our method is still significant. The classification performance of 1D-CNN is more superior than that of SVM in OA, AA and Kappa, but the classification performance of our method is much better than that of SVM and 1D-CNN. Compared with 1D-CNN, the classification accuracy of Forest has the largest improvement, increased from 78.73\% to 99.94\%. The classification accuracy of all categories has been improved to varying degrees, and the classification accuracy of Potato, Wheat, Water and Pea increased by more than 10\%. As a result, OA is improved from 87.46\% to 98.97\%, AA is improved from 89.02\% to 98.62\%, and the Kappa coefficient is improved from 0.86 to 0.99.
 


\section{Conclusion}
This paper proposed a novel method for the land cover classification of remote sensing imagery, which introduced the coordinate information to enhance the expression of pixel features. Dual-branch networks are designed to learn spatial feature and pixel feature, respectively. The features of two branches are fused by addition and the classification task is realized by another FCN. Finally, two experiments have been conducted to prove the effectiveness of the improved method with two kinds of remote sensing datasets, involving hyperspectral image and PolSAR image.

\bibliographystyle{IEEEtran}
\bibliography{nj}

\begin{thebibliography}{10}
\providecommand{\url}[1]{#1}
\csname url@samestyle\endcsname
\providecommand{\newblock}{\relax}
\providecommand{\bibinfo}[2]{#2}
\providecommand{\BIBentrySTDinterwordspacing}{\spaceskip=0pt\relax}
\providecommand{\BIBentryALTinterwordstretchfactor}{4}
\providecommand{\BIBentryALTinterwordspacing}{\spaceskip=\fontdimen2\font plus
\BIBentryALTinterwordstretchfactor\fontdimen3\font minus
  \fontdimen4\font\relax}
\providecommand{\BIBforeignlanguage}[2]{{%
\expandafter\ifx\csname l@#1\endcsname\relax
\typeout{** WARNING: IEEEtran.bst: No hyphenation pattern has been}%
\typeout{** loaded for the language `#1'. Using the pattern for}%
\typeout{** the default language instead.}%
\else
\language=\csname l@#1\endcsname
\fi
#2}}
\providecommand{\BIBdecl}{\relax}
\BIBdecl

\bibitem{Douglas2015Mapping}
D.~G. Goodin, K.~L. Anibas, and M.~Bezymennyi, ``Mapping land cover and land
  use from object-based classification: an example from a complex agricultural
  landscape,'' \emph{Remote Sensing}, vol.~36, no.~18, pp. 4702--4723, 2015.

\bibitem{TONG2020111322}
X.-Y. Tong, G.-S. Xia, Q.~Lu, H.~Shen, S.~Li, S.~You, and L.~Zhang,
  ``Land-cover classification with high-resolution remote sensing images using
  transferable deep models,'' \emph{Remote Sensing of Environment}, vol. 237,
  p. 111322, 2020.

\bibitem{Bhosle2019}
K.~Bhosle and V.~Musande, ``Evaluation of deep learning cnn model for land use
  land cover classification and crop identification using hyperspectral remote
  sensing images,'' \emph{Journal of the Indian Society of Remote Sensing},
  vol.~47, pp. 1949--1958, 2019.

\bibitem{Ma2015Hyperspectral}
X.~Ma, J.~Geng, and H.~Wang, ``Hyperspectral image classification via
  contextual deep learning,'' \emph{Eurasip Journal on Image \& Video
  Processing}, vol. 2015, no.~1, p.~20, 2015.

\bibitem{zhang2019sar}
F.~Zhang, Y.~Wang, J.~Ni, Y.~Zhou, and W.~Hu, ``Sar target small sample
  recognition based on cnn cascaded features and adaboost rotation forest,''
  \emph{IEEE Geoscience and Remote Sensing Letters}, pp. 1--5, 2019.

\bibitem{8921180}
Y.~{Song}, Z.~{Zhang}, R.~K. {Baghbaderani}, F.~{Wang}, Y.~{Qu}, C.~{Stuttsy},
  and H.~{Qi}, ``Land cover classification for satellite images through 1d
  cnn,'' in \emph{2019 10th Workshop on Hyperspectral Imaging and Signal
  Processing: Evolution in Remote Sensing (WHISPERS)}, 2019, pp. 1--5.

\bibitem{Okeke}
O.~Stephen, O.~Stephen, S.~Ibrokhimov, and K.~L. Hui, ``A multiple-loss
  dual-output convolutional neural network for fashion class classification,''
  in \emph{2019 21st International Conference on Advanced Communication
  Technology (ICACT)}, 2019.

\bibitem{zhong2017a}
P.~Zhong and Z.~Gong, ``A hybrid dbn and crf model for spectral-spatial
  classification of hyperspectral images,'' \emph{Statistics, Optimization and
  Information Computing}, vol.~5, no.~2, pp. 75--98, 2017.

\bibitem{Cao2018Hyperspectral}
X.~Cao, F.~Zhou, L.~Xu, D.~Meng, Z.~Xu, and J.~Paisley, ``Hyperspectral image
  classification with markov random fields and a convolutional neural
  network,'' \emph{IEEE Transactions on Image Processing A Publication of the
  IEEE Signal Processing Society}, pp. 1--1, 2018.

\bibitem{8127432}
C.~{Danilla}, C.~{Persello}, V.~{Tolpekin}, and J.~R. {Bergado},
  ``Classification of multitemporal sar images using convolutional neural
  networks and markov random fields,'' in \emph{2017 IEEE International
  Geoscience and Remote Sensing Symposium (IGARSS)}, 2017, pp. 2231--2234.

\bibitem{zhongyang2019terrain}
Z.~Zhongyang, C.~Yinglei, S.~Xiaosong, Q.~Xianxiang, and L.~Xin, ``Terrain
  classification of lidar point cloud based on multi-scale features and
  pointnet,'' \emph{Laser \& Optoelectronics Progress}, vol.~56, no.~5, p.
  052804, 2019.

\bibitem{wolf2015fast}
D.~Wolf, J.~Prankl, and M.~Vincze, ``Fast semantic segmentation of 3d point
  clouds using a dense crf with learned parameters,'' in \emph{international
  conference on robotics and automation}, 2015, pp. 4867--4873.

\bibitem{krahenbuhl2011efficient}
P.~Krahenbuhl and V.~Koltun, ``Efficient inference in fully connected crfs with
  gaussian edge potentials,'' in \emph{international conference on robotics and
  automation}, 2011, pp. 109--117.

\bibitem{Chen2014Deep}
Y.~Chen, Z.~Lin, Z.~Xing, W.~Gang, and Y.~Gu, ``Deep learning-based
  classification of hyperspectral data,'' \emph{IEEE Journal of Selected Topics
  in Applied Earth Observations \& Remote Sensing}, vol.~7, no.~6, pp.
  2094--2107, 2014.

\bibitem{Liu2018An}
T.~Liu and A.~E. Amr, ``An object-based image analysis method for enhancing
  classification of land covers using fully convolutional networks and
  multi-view images of small unmanned aerial system,'' \emph{Remote Sensing},
  vol.~10, no.~3, p. 457, 2018.

\bibitem{zhang2017nearest}
F.~Zhang, J.~Ni, Q.~Yin, W.~Li, Z.~Li, Y.~Liu, and W.~Hong,
  ``Nearest-regularized subspace classification for polsar imagery using
  polarimetric feature vector and spatial information,'' \emph{Remote Sensing},
  vol.~9, no.~11, p. 1114, 2017.

\bibitem{xiong2014representation}
M.~Xiong, F.~Zhang, Q.~Ran, W.~Hu, and W.~Li, ``Representation-based
  classifications with markov random field model for hyperspectral urban
  data,'' \emph{Journal of Applied Remote Sensing}, vol.~8, no.~1, p. 085097,
  2014.

\bibitem{teichmann2018convolutional}
M.~T. Teichmann and R.~Cipolla, ``Convolutional crfs for semantic
  segmentation,'' \emph{arXiv preprint arXiv:1805.04777}, 2018.

\bibitem{8726302}
Z.~{Zhong}, J.~{Li}, D.~A. {Clausi}, and A.~{Wong}, ``Generative adversarial
  networks and conditional random fields for hyperspectral image
  classification,'' \emph{IEEE Transactions on Cybernetics}, pp. 1--12, 2019.

\bibitem{samiappan2016fusion}
S.~Samiappan, L.~Dabbiru, and R.~Moorhead, ``Fusion of hyperspectral and lidar
  data using random feature selection and morphological attribute profiles,''
  in \emph{2016 8th Workshop on Hyperspectral Image and Signal Processing:
  Evolution in Remote Sensing (WHISPERS)}.\hskip 1em plus 0.5em minus
  0.4em\relax IEEE, 2016, pp. 1--4.

\bibitem{blanco2010synergistic}
P.~Blanco, H.~del Valle, P.~Bouza, G.~Metternicht, and A.~Zinck, ``Synergistic
  use of landsat and hyperion imageries for ecological site classification in
  rangelands,'' in \emph{2010 2nd Workshop on Hyperspectral Image and Signal
  Processing: Evolution in Remote Sensing}.\hskip 1em plus 0.5em minus
  0.4em\relax IEEE, 2010, pp. 1--4.

\bibitem{wang2017research}
H.~Wang, J.~Zhang, J.~Wu, and Z.~Yan, ``Research on mangrove recognition based
  on hyperspectral unmixing,'' in \emph{2017 IEEE International Conference on
  Unmanned Systems (ICUS)}.\hskip 1em plus 0.5em minus 0.4em\relax IEEE, 2017,
  pp. 298--300.

\bibitem{hu2015deep}
W.~Hu, Y.~Huang, L.~Wei, F.~Zhang, and H.~Li, ``Deep convolutional neural
  networks for hyperspectral image classification,'' \emph{Journal of Sensors},
  vol. 2015, no. 2015, pp. 1--12, 2015.

\bibitem{ahishali2020multifrequency}
M.~Ahishali, S.~Kiranyaz, T.~Ince, and M.~Gabbouj, ``Multifrequency polsar
  image classification using dual-band 1d convolutional neural networks,'' in
  \emph{2020 Mediterranean and Middle-East Geoscience and Remote Sensing
  Symposium (M2GARSS)}.\hskip 1em plus 0.5em minus 0.4em\relax IEEE, 2020, pp.
  73--76.

\end{thebibliography}

\end{document}